\documentclass[a4paper,fleqn]{cas-dc}

\usepackage[authoryear]{natbib}
\usepackage{amsmath}
\usepackage{bm}
\usepackage{soul}
\usepackage{makecell}
\usepackage[utf8]{inputenc}
\DeclareUnicodeCharacter{2606}{\textsuperscript{*}}

\begin{document}
\let\WriteBookmarks\relax
\def\floatpagepagefraction{1}
\def\textpagefraction{.001}

\shorttitle{}    

\shortauthors{Haoda Li et al.}  

\title [mode = title]{Control System Design and Experiments for Autonomous Underwater Helicopter Docking Procedure Based on Acoustic-inertial-optical Guidance}                      


\affiliation[1]{organization={Ocean college,Zhejiang University},
    city={Zhoushan},
    postcode={316021}, 
    country={China}}
\affiliation[2]{organization={Hainan Institute,Zhejiang University},
    city={Sanya},
    postcode={572025}, 
    country={China}}
    
\author[1]{Haoda Li}[
orcid=0009-0006-2274-3077,
]
\credit{Conceptualization of this study, Software,Data curation, Validation, Formal analysis, Writing – original draft,  Writing – review & editing}

\author[1]{Xinyu An}[style=chinese]
\credit{Conceptualization of this study, Software, Validation, Writing – review, Funding acquisition}

\author[1,2]{Rendong Feng}[]
\credit{Conceptualization of this study, Software, Validation, Writing – review}

\author[1,2]{Zhenwei Rong}[]
\credit{Software, Validation}

\author[1]{Zhuoyu Zhang}[]
\credit{Conceptualization of this study, Software, Validation}

\author[1]{Zhipeng Li}[]
\credit{Conceptualization of this study, Software, Validation}

\author[1]{Liming Zhao}[]
\credit{Validation, Software}


\author[1]{Ying Chen}
\credit{Writing – review, Funding acquisition}

\cormark[1]
\nonumnote{* Corresponding author. \newline \indent Email address: ychen@zju.edu.cn(Ying Chen).}

\begin{abstract}
A control system structure for the underwater docking procedure of an Autonomous Underwater Helicopter (AUH) is proposed in this paper, which utilizes acoustic-inertial-optical guidance. Unlike conventional Autonomous Underwater Vehicles (AUVs), the maneuverability requirements for AUHs are more stringent during the docking procedure, requiring it to remain stationary or have minimal horizontal movement while moving vertically. The docking procedure is divided into two stages: Homing and Landing, each stage utilizing different guidance methods. Additionally, a segmented aligning strategy operating at various altitudes and a linear velocity decision are both adopted in Landing stage. Due to the unique structure of the Subsea Docking System (SDS), the AUH is required to dock onto the SDS in a fixed orientation with specific attitude and altitude. Therefore, a particular criterion is proposed to determine whether the AUH has successfully docked onto the SDS. Furthermore, the effectiveness and robustness of the proposed control method in AUH's docking procedure are demonstrated through pool experiments and sea trials.
\end{abstract}


\maketitle

\section{Introduction}
The field of underwater technology has experienced a significant development in both technical and economic aspects \citep{Trslic2020}, especially in the realm of underwater robotics. Over the past few decades, ocean observation and other applications has expanded from shallow sea to deep sea, from surface water to the ocean floor, drawing greater attention to subsea robots. These robots plays crucial roles in many tasks, including inspecting pipelines, seabed mineral exploration, tracking cable routes and ocean observation \citep{Rumson2021, Yu2018}.

MBARI Rover II \citep{Jr2021,Mcgill2007} is an autonomous dual-track driven submarine exploration equipment. It can measure sea water temperature, oxygen concentration, flow rate and sediment community oxygen consumption by using the onboard sensor. It can cross the sea floor under low ground contact pressure, take pictures of the sea floor conditions, and stay regularly to allow the respiration meter to hatch and measure carbon remineralization. The FlatFish is a compact autonomous underwater vehicle(AUV), designed to acquire a high-resolution, textured 3D model of an underwater structure within an oil and gas asset \citep{Albiez2016}. Similarly, the Eelume is designed to reside subsea to provide immediate response to unpredictable inspection, maintenance and repair (IMR) requirements \citep{Liljeback2017}.

\begin{figure}[h]
	\centering
\includegraphics[width=0.48\textwidth]{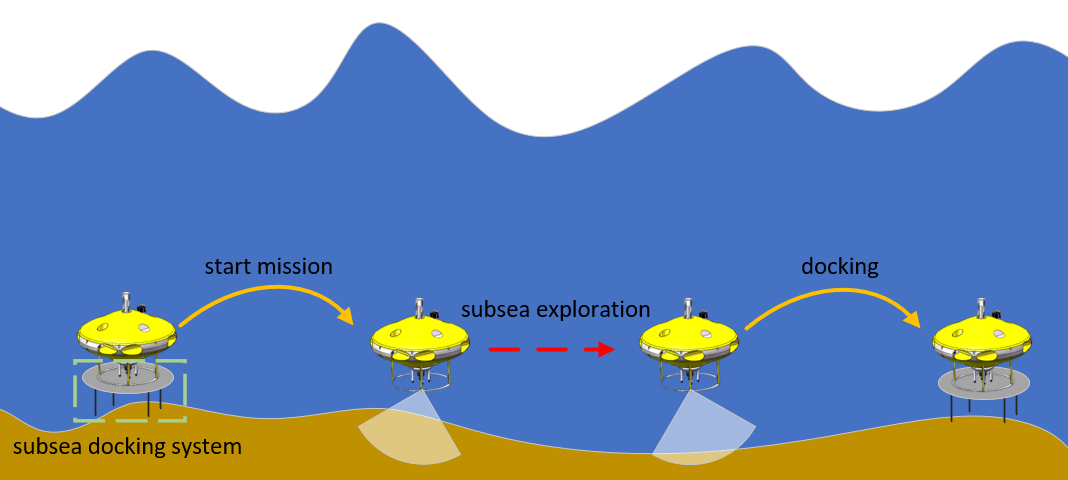}
	\caption{Operating pattern of the AUH near seafloor. \citep{Li2023}.}
	\label{FIG:1.1}
\end{figure}

However,the aforementioned subsea robots all share a common drawback, namely a lack of maneuverability. \cite{Zhou2023} explained that the concept of maneuverability refers to the ability to change speed, direction, and location while maintaining stability. An autonomous underwater helicopter (AUH), a disk type of AUV dedicated to subsea operations, provides several advantages, including seabed exploration, fixed-point hovering, high maneuverability, anti-flow stability, etc \citep{Wang2019, Zhou2022}. Nonetheless, compared with typical torpedo-type AUVs, the AUH exhibits poor hydrodynamic performance and a higher resistance coefficient, resulting in limited endurance. To address this issue and enhance the endurance of AUVs, researchers have proposed a subsea docking system (SDS). This system allow AUVs to recharge, upload data and upgrade mission by docking onto the SDS \citep{Singh2001, Li2015, Guo2006a}. Inspired by previous research, Cai \citep{cai2023resident} proposed a resident subsea docking system specifically designed for the AUH. By incorporating the SDS, the AUH can perform a series of tasks, including communication, equipment maintenance, charging, and more, as it travels between subsea stations (shown in Figure \ref{FIG:1.1}).

\begin{table*}[]
\caption{Comparison of various guidance methods. \citep{Palmer2009, Evans2003, Stokey2001}}\label{various guidance}
\begin{tabular}{llll}
\toprule
guidance principle        & sensors             & advantage    & disadvantage  \\            \midrule
acoustic-inertial-optical & DVL, IMU, camera      & high accuracy & \makecell[l]{for short range navigation,\\ within the last ten meters} \\
acoustic-inertial         & USBL, IMU            & for large range navigation         & accuracy on the order of meters \\
acoustic-optical          & sonar, camera       & high accuracy       &\makecell[l]{ navigation range is depended on sonar \\(about 100m)}  \\
acoustic-inertial-optical & USBL, DVL, IMU,camera & high accuracy and large range navigation  & expensive and take up more space  \\                                                         \bottomrule     
\end{tabular}
\end{table*}

In present, most docking procedure are carried out using torpedo-type AUVs with varying guidance methods. \cite{Park2009a} presented an underwater docking procedure for the test-bed autonomous AUV platform called ISiMI and proposed a final approach algorithm based on optical guidance. \cite{Li2015} proposed a homing control methods for USBL-optical navigation and the docking system has been tested in a water pool. Similar work has been done by \cite{Allen}, \citep{Singh2001} and \citep{McEwen2008}. Unlike conventional docking procedure with torpedo-type AUVs, AUH demand for a higher spatial mobility while torpedo-type AUVs emphasize mobility in the vertical plane. Table \ref{various guidance} lists a comparison of various guidance methods. After comprehensive evaluation, the acoustic-inertial-optical guidance, which is based on an ultra-short baseline (USBL) positioning system, a doppler velocity log (DVL), an inertial measurement unit (IMU) and a monocular camera, is adopted to the control system. This study focus on the control system applied in docking procedure and structure of the AUH, including software and hardware. Pool experiments and sea trial were conducted in South China Sea.

The rest of this paper is organized as follows. Section \ref{Components of the Docking procedure} introduces the key components of the docking procedure, including the AUH, the SDS, and the influence factors of docking procedure. Section \ref{System architecture of AUH} presents the system architecture of AUH, both in hardware and software, and introduces the subsystem of motion control. Section \ref{Guidance and control method} describes an overview of guidance method and emphasizes the control flow. Besides, a criterion will be proposed to determine whether the AUH successfully dock onto the SDS. Pool experiments and sea trial  will be introduced and discussed in section \ref{Experiments}, and conclusions will be given in section \ref{conclusions}.

\section{Components of the docking procedure}\label{Components of the Docking procedure}
\begin{figure}[h]
	\centering
\includegraphics[width=0.48\textwidth]{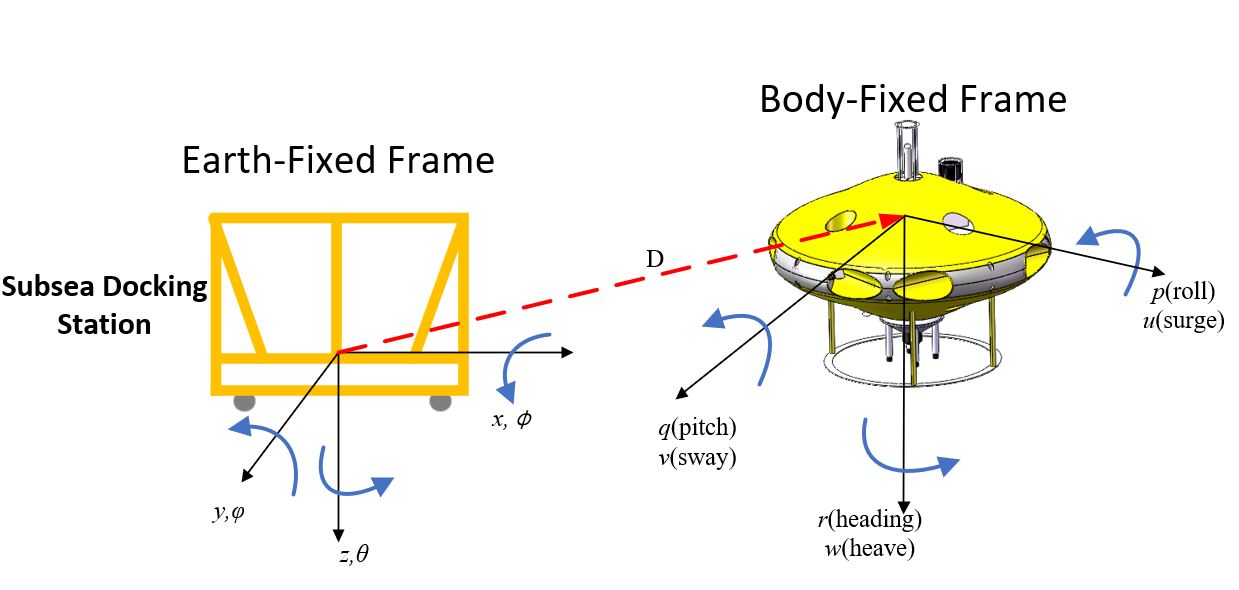}
	\caption{The coordinate of the AUH.}
	\label{FIG:2.1}
\end{figure}

The definition of the coordinate system is shown in Fig. \ref{FIG:2.1}. The earth-fixed frame is also known as the inertial coordinate system, whose original is fixed on the center of SDS. Meanwhile, the original of body-fixed frame is located at the center of buoyancy of the AUH. 

In Fig. \ref{FIG:2.1}, $\bm{\xi}  = {[x,y,z]^T}$ represents the absolute linear displacement in the inertial coordinate system while $\bm{\eta}  = {[\phi ,\varphi ,\theta ]^T}$ represents angular position \citep{fossen2011handbook}. The diameter and height of AUH is 2 m and 1.6 m, with a weight of 760 kg in air given in Table \ref{specification of AUH}. The geometrical size (length * width * height) of SDS is 3 m * 3 m * 2 m, with a vertical funnel-shaped docking entrance inside \citep{cai2023resident}.

Multiple factors impact the success rate of a docking procedure:
\begin{itemize}
\item Accuracy of guidance method.
\item Maneuverability of the AUH.
\item Performance of motion control.
\end{itemize}

The success of AUH docking primarily relies on a high-precision guidance method, which determines whether the AUH can accurately return to the base station from a distance and descend onto the SDS from a particular altitude. Once the precision of the guidance method is adequately assured, consideration must be given to the AUH's motion performance. During the descent from a specific altitude, it is essential to maintain a constant horizontal position, which requires the controller with sufficient response speed and high steady-state accuracy for efficient docking. Furthermore, due to the design requirements of AUH, a specific attitude and altitude must be maintained during the docking procedure. Consequently, the stationary hovering and full circle turning capabilities of AUH can be fully utilized.

\begin{table}[!h]
\caption{Main specifications of AUH}\label{specification of AUH}
\begin{tabular}{@{}ll@{}}
\toprule
Features        &Description                                      \\
\midrule
Diameter        & 2 m                                                                                                                   \\ 
Height          & 1.6 m                                                                                                                 \\
Weight in air   & 760 kg                                                                                                                \\
Speed           & 1 m/s in normal                                                                                                     \\
Operating depth & 0$\sim$1500 m                                                                                                         \\
Component       & \makecell[l]{inertial measurement unit(IMU), depth \\altimeter, monocular camera, ultra-short \\baseline (USBL) positioning system\\ doppler velocity log (DVL), acoustic\\ transmitter}\\
\bottomrule
\end{tabular}
\end{table}

\section{System architecture of AUH}\label{System architecture of AUH}

\begin{figure}[!h]
	\centering
\includegraphics[width=0.48\textwidth]{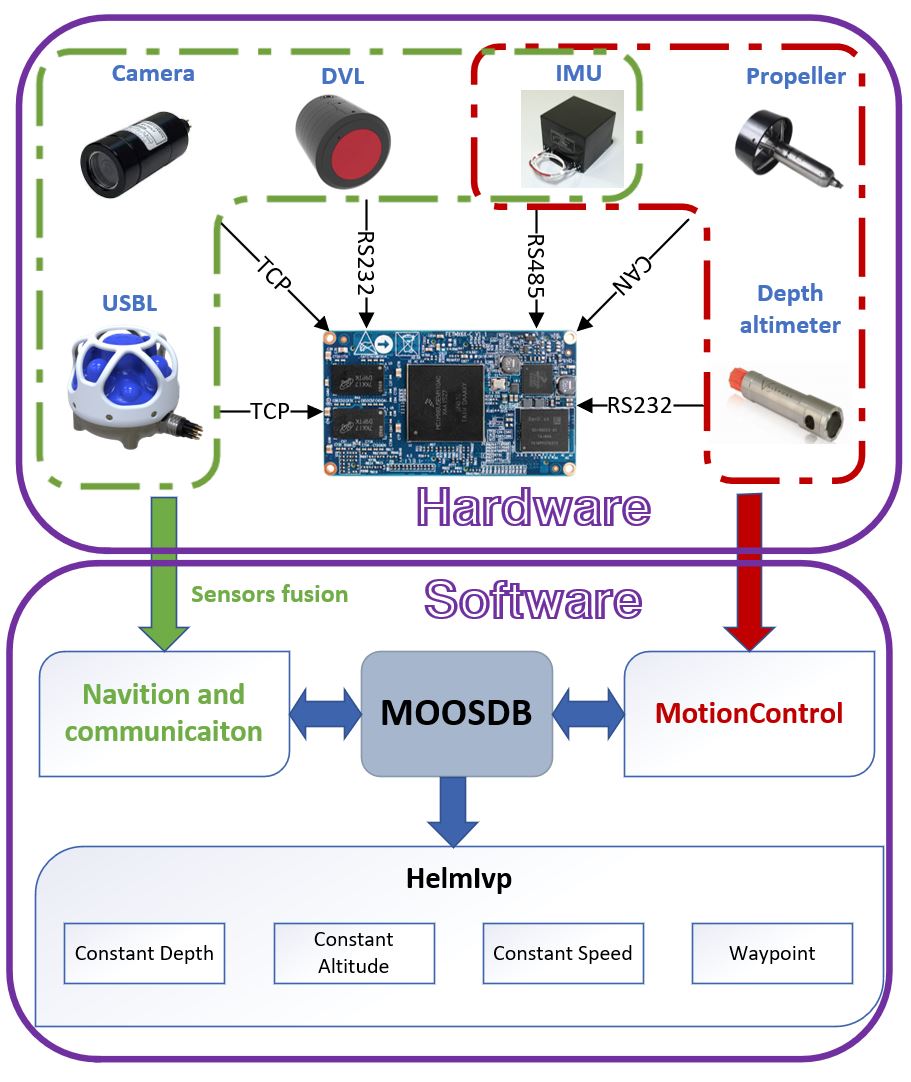}
	\caption{Hardware and software structure of the AUH.}
	\label{FIG:2.2}
\end{figure}

This section introduces the hardware and software architecture of the AUH, as illustrated in Figure \ref{FIG:2.2}. The hardware consists of main control panel, USBL, camera, DVL, IMU, propellers and depth altimeter. Additionally, various communication protocols, such as RS232, RS485, CAN, TCP are used to collect data and send it to the main control panel. In terms of software, the control system of AUH is developed based on open-source MOOS-Ivp, which is a set of C++ modules for providing autonomy on robotic platforms especially autonomous marine vehicles. The control system of the AUH includes four essential modules: MOOSDB, Navigation and Communication, Motion Control and HelmIvp. There are four basic behaviours, namely Constant Depth, Constant Altitude, Constant Speed and Waypoint, that can be executed by AUH and managed by HelmIvp module. These four behaviours are decoupled and can be combined with each other except for Constant Depth and Constant Altitude. Furthermore, the Waypoint behavior indicates that the AUH moves from one location to another while maintaining a yaw angle.

\begin{figure*}[!h]
	\centering
\includegraphics[width=0.95\textwidth]{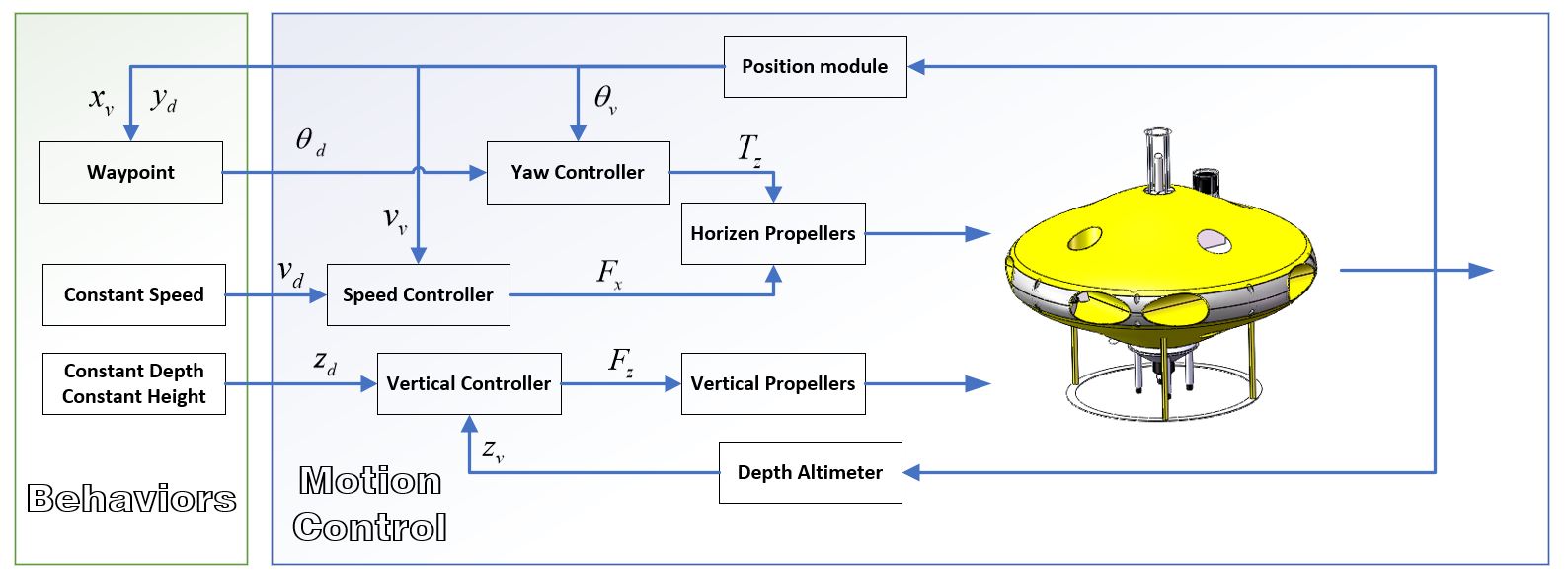}
	\caption{The subsystem of motion control.}
	\label{FIG:2.3}
\end{figure*}

In Figure \ref{FIG:2.3}, $x_v$, $y_v$, $z_v$, $\theta_v$ and $v_v$ respectively refer to current three-dimension position, yaw angle and speed of AUH, which can be obtained by IMU, depth altimeter and DVL. $\theta_d$, $v_d$ and $z_d$ respectively represent desired values of yaw, speed and depth. As shown as Figure \ref{FIG:2.3}, four behaviours generate desired values and deliver them into controllers, then controllers calculate the final torque $T_z$, horizontal thrust $F_x$, and vertical thrust $F_z$. In addition, yaw controller, speed controller and vertical controller are all based on proportional-integration-differential (PID) principle. However, PID controller is known for its poor robustness and weak anti-interference. To address these concerns, the linear active disturbance rejection control with tracking differentiator (LADRC-TD) principle will replace PID in the yaw controller and depth controller in future work.

\section{Guidance method and docking stages}\label{Guidance and control method}
\subsection{Guidance method}
The docking procedure is divided into two stages: Homing and Landing, each with its own guidance methods. 
\begin{itemize}
    \item Homing. Homing is when the AUH returns to the SDS from a distance after completing a task or when the battery is running low. In this stage, multi-sensor fusion navigation technology based on acoustic-inertial principle is used as the guidance method
    \item Landing. Landing commences after the Homing stage. The AUH descends slowly towards the center of the SDS in a specific orientation with precise attitude and altitude, and this process is called Landing. While multi-sensor fusion navigation technology can guide the AUH back to the SDS, its accuracy is inadequate for the docking process because of errors on the order of meters. This inaccuracy greatly impacts the docking success rate. To solve this issue, an optical-guidance system is adopted in the Landing stage, which is fast, robust, and easy to operate \citep{Deltheil2000}.
\end{itemize}

During the Homing stage, a multi-sensor navigation system based on acoustic-inertial principle is utilized. This system, which is a commercial product, consists of IMU, DVL and USBL. However, the detailed description of this system is beyond the scope of this article. Unlike the acoustic-inertial guidance method in Homing stage, the visual-guidance system used in Landing stage is based on a monocular camera and its principle is introduced below.

\begin{figure}[!h]
	\centering
\includegraphics[width=0.4\textwidth]{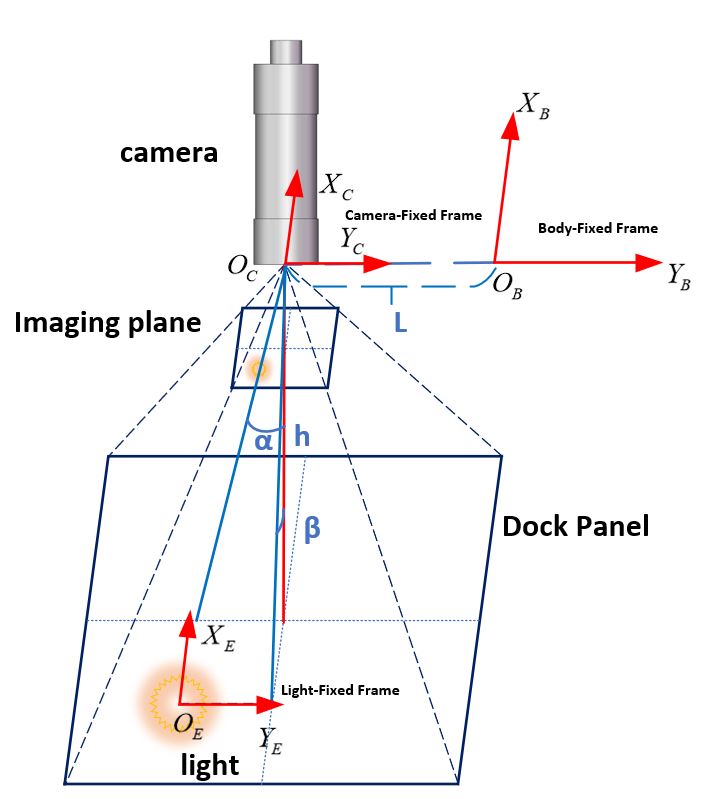}
	\caption{The coordinate of the optical imaging system.}
	\label{FIG:3.3}
\end{figure}

As depicted in Fig. \ref{FIG:3.3}, the light first generates a spot on the imaging plane. The geometric center of maximum spot should be then calculated, along with the equivalent deviation angles in the horizontal $(\alpha)$ and vertical plane $(\beta)$. \citep{Lin2022, Lin2019}
\begin{equation}
\begin{split}
\noindent \alpha  =& \arctan (\frac{{2\bar u}}{M} \times \tan {\alpha _0})\\
\noindent \beta  =& \arctan (\frac{{2\bar v}}{N} \times \tan {\beta _0})
\end{split}
\label{equ 1}
\end{equation}
where, $(\bar u,\bar v)$ denotes the center location of the maximum spot  on the image centered at the optic center of camera, $M \times N$ represents the pixels of the image. $\alpha_0$ and $\beta_0$ refer to the field angles of the camera in the horizontal and vertical plane, respectively. Combined with $\alpha$ and $\beta$, the coordinate of light in camera-fixed frame $(x_c^l,y_c^l)^T$ can be described as
 
\begin{flalign}
\begin{split}
& x_C^l = h \times \tan \beta \\
& y_C^l = h \times \tan \alpha 
\label{equ 2}
\end{split}
\end{flalign}
where, $h$ denotes the difference in depth between the light located on SDS and the camera installed in AUH. Furthermore, the camera is not fixed on the center of AUH and the distance between them is L, as shown in Fig. \ref{FIG:3.3}. 
\begin{align}
\begin{split}
 x_B^l =& h \times \tan \beta \\
 y_B^l =& h \times \tan \alpha  - L
\end{split}
\label{equ 3}
\end{align}
where, $(x_B^l,y_B^l)^T$ represent the coordinate of the light in the body-fixed frame of reference. To obtain the coordinates of the light in the earth-fixed frame of reference relative to the AUH, a transformation matrix is applied. The transformation process is carried out as follows:

\begin{flalign}
\begin{split}
\left[{\begin{array}{*{20}{l}}
 x\\
 y
\end{array}} \right] =&  - \left[ {\begin{array}{*{20}{l}}
{\cos \theta }&{ - \sin \theta }\\
{\sin \theta }&{\cos \theta }
\end{array}} \right]\left[ {\begin{array}{*{20}{l}}
{x_B^l}\\
{y_B^l}
\end{array}} \right]\\
 = &\left[ {\begin{array}{*{20}{l}}
{h \cdot \tan \alpha  \cdot \sin \theta  - h \cdot \tan \beta  \cdot \cos \theta  - L \cdot \sin \theta }\\
{ - h \cdot \tan \beta  \cdot \sin \theta  - h \cdot \tan \alpha  \cdot \cos \theta  + L \cdot \cos \theta }
\end{array}} \right]
\label{equ 4}
\end{split}
\end{flalign}

where, $(x,y)^T$ refers to the coordinate of AUH in the earth-fixed frame.

\subsection{Homing stage}
\begin{figure*}[!h]
	\centering
\includegraphics[width=0.9\textwidth]{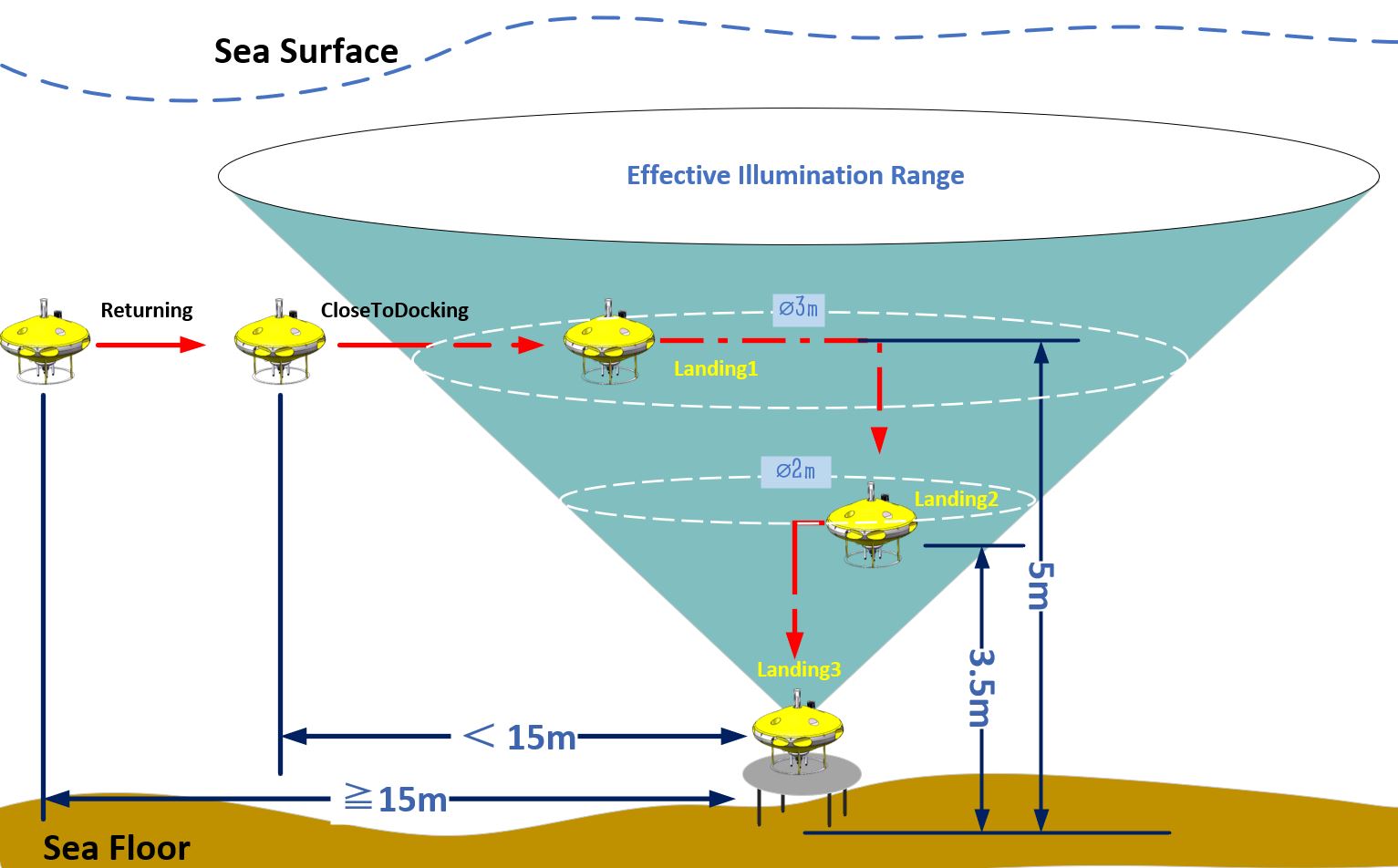}
	\caption{Segmented docking procedure.}
	\label{FIG:3.1}
\end{figure*}

\begin{figure*}[htp]
	\centering
\includegraphics[width=0.95\textwidth]{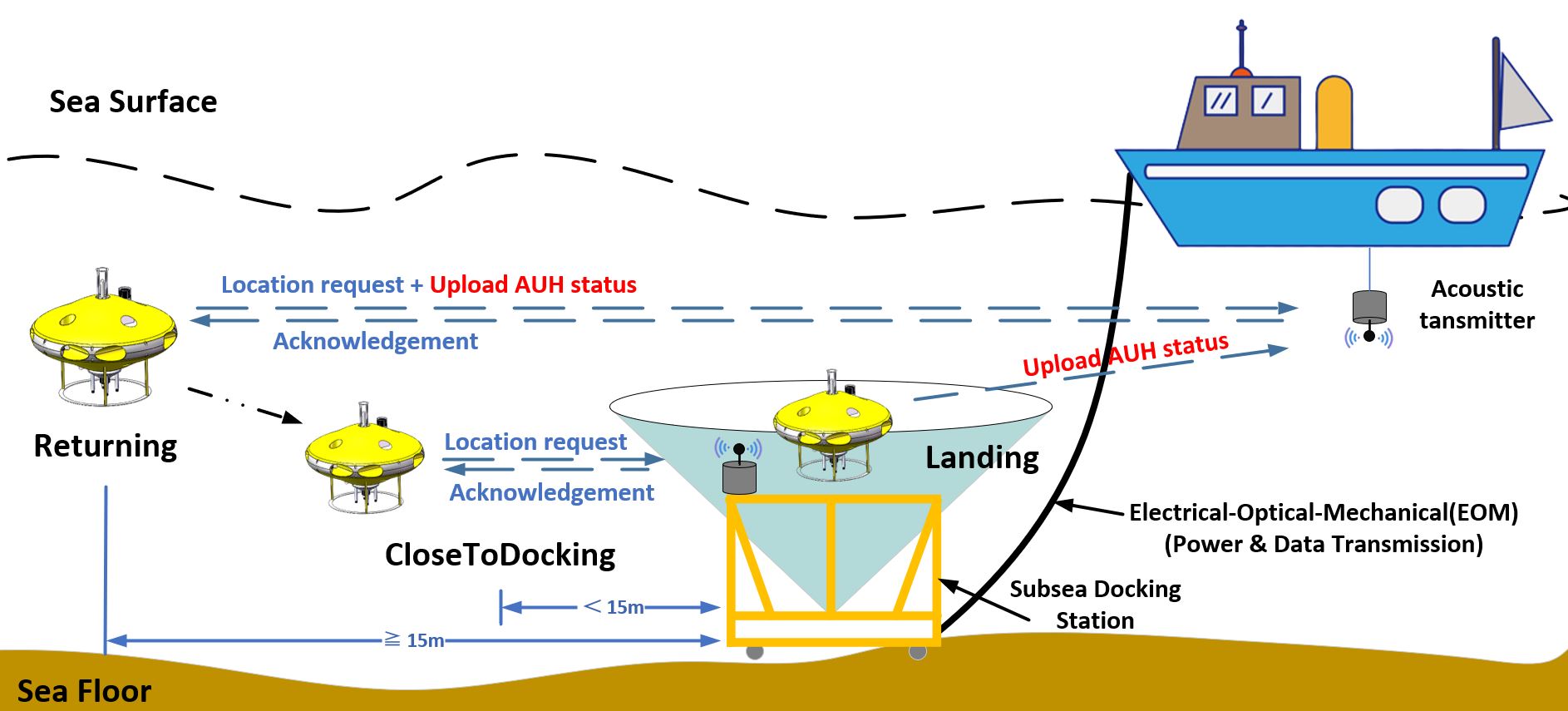}
	\caption{The navigation and communication in Returning and CloseToDocking phase.}
	\label{FIG:3.2}
\end{figure*}

The Homing stage can be divided into two phases: Returning and CloseToDocking (as shown in Fig. \ref{FIG:3.1}). In the Returning phase, AUH cruises back to the SDS using the acoustic-inertial navigation system until it is 15 meters from the SDS. During this phase, the acoustic transmitter on the AUH serves dual purposes: it works as part of the USBL location system and also operates acoustic communication. The AUH can upload 120 bytes of operating status data to the system monitor mounted on the ship (as shown in Fig. \ref{FIG:3.2}). Typically, it takes around 20 second for the acoustic transmitter to upload the data. Considering the time required for location request and acknowledgement, the USBL system operates at a frequency of up to 1.5 times per minute. Generally, a higher USBL frequency contributes to greater precision acoustic-inertial fusion navigation system, enabling the AUH to approach the SDS more efficiently. The phase transition from Returning to CloseToDocking when AUH reaches a distance of 15 meters or less from the SDS. In the CloseToDocking phase, the USBL frequency increases to 3 times per minute, and the acoustic transmitter only sends a location request without uploading status information. The primary objective in this phase is to enhance the precision of the acoustic-inertial guidance method.

It is important to note that the cruising speed differs between the Returning phase and the CloseToDocking phase. In the Returning phase, the AUH cruises back to the SDS at a speed of 1 m/s in order to minimize the time taken. However, when the AUH gets closer to the SDS, cruising at 1 m/s can cause the AUH to be easily deflected out of the effective illumination range. Therefore, in the CloseToDocking phase, it is more suitable for the AUH to cruise at a slower speed of 0.3 m/s to ensure it remains within the effective illumination range.
In terms of vertical motion, the AUH primarily operates using the Constant Altitude behavior. However, during the CloseToDocking phase, the altitude value measured by the depth altimeter, which is based on acoustical principles, fluctuates significantly. This is because the sound waves may be obstructed by the mental structure of the SDS when the AUH is directly above it. Hence, in the CloseToDocking phase, the AUH operates using the Constant Depth behavior, which differs from the Returning phase.(as shown in Table \ref{table 1})

\begin{table}[width=1\linewidth,cols=3,pos=h]
\caption{The parameters of Returning and CloseToDocking.}\label{tbl1}
\begin{tabular*}{\tblwidth}{@{} LLL@{} }
\toprule
   & Returning & CloseToDocking\\
\midrule
range(m)          & \textgreater{}=15    & \textless{}15          \\
speed(m/s)        & 1                    & 0.3                    \\
vertical motion   & Constant Altitude      & Constant Depth             \\
navigation mode   & IMU+DVL+USBL         & IMU+DVL+USBL           \\
USBL's frequency & 1.5 times per minute & 3 times per minute     \\
upload data       & Yes                  & No                   \\
\bottomrule
\label{table 1}
\end{tabular*}
\end{table}

\subsection{Landing stage}

When the AUH approaches the SDS within 15 m and enters the effective range of illumination, it transitions from the Returning phase to the CloseToDocking phase. The visible range of the illumination forms a cone-shaped area in space, which is related to AUH's working altitude. The calculation for this visible range is presented below:
\begin{equation}
\begin{split}
R  =& \tan\frac{\rho}{2} \times h
\end{split}
\label{equ radius}
\end{equation}

where, $R$ denotes the effective illuminate radius and $h$ represents the AUH's working altitude. $\rho$ refers to the \sethlcolor{green}\hl{divergence angle} of the camera and it's value in this paper is $70^{\circ}$. During the actual implementation, the working altitude in Landing1 and Landing2 is 5 m and 3.5 m, respectively. As a consequence, their corresponding effective illuminate radius is about 3 m and 2 m.

\begin{figure}[htp]
	\centering
\includegraphics[width=0.4\textwidth]{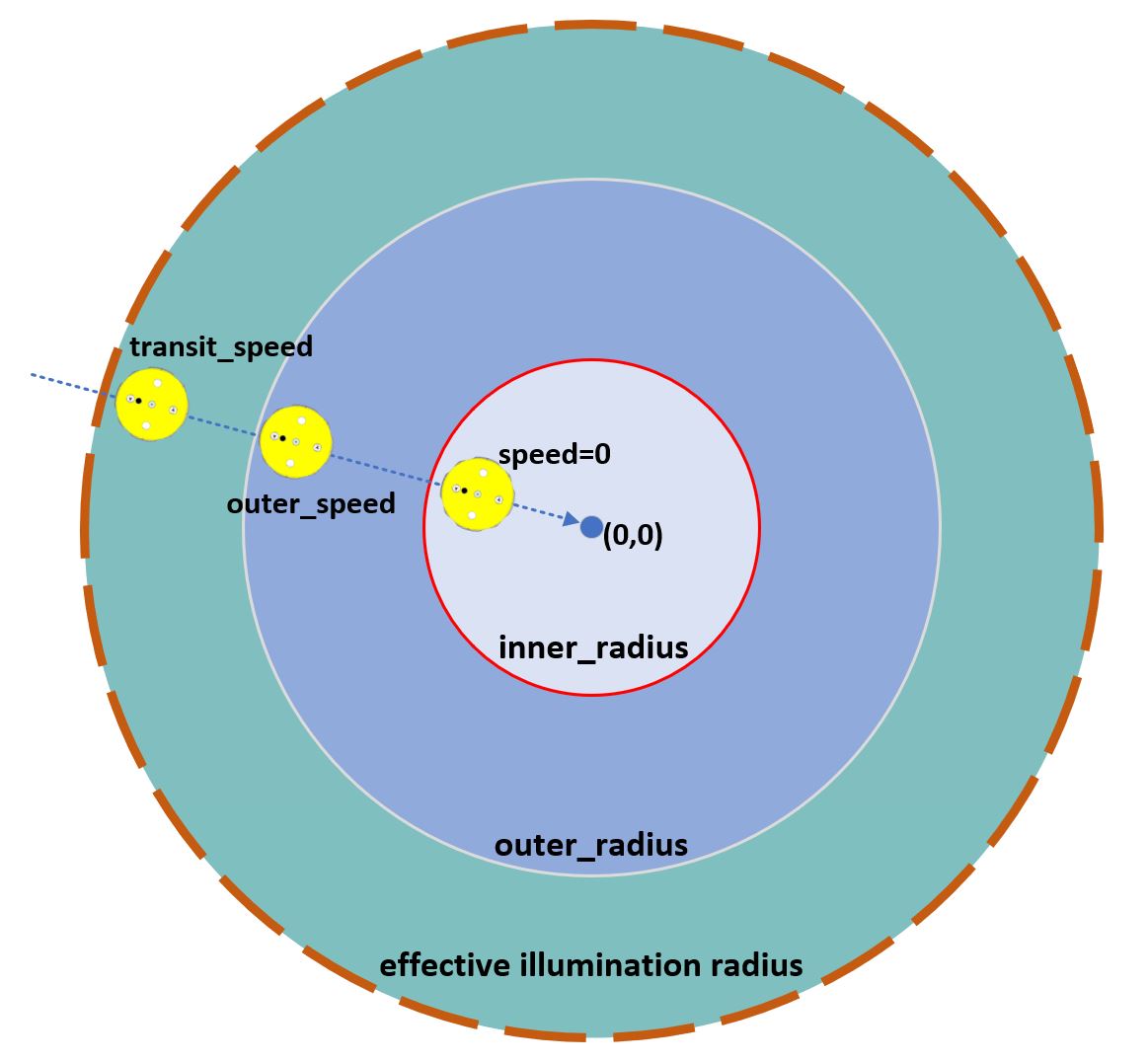}
	\caption{Schematic diagram of speed decision strategy.}
	\label{FIG:3.6}
\end{figure}

During the descent from a certain altitude while trying to maintain the AUH within the effective illumination range, the AUH encounters the difficulties caused by the influence of inertia and current disturbances. To address this, a segmented aligning strategy is employed in the Landing stage, which is divided into three phases: Landing1, Landing2, Landing3, and each of these phases operates at various altitudes. The parameters for each phase are detailed in Table \ref{table 2}. \sethlcolor{green}\hl{The distance threshold for each phase is an empirical value that can be adjusted based on changing environmental conditions. In the "work altitude" column, the three phases operate at different altitudes: 5 m, 3.5 m, and 0.2 m. Additionally, the altitude of Landing3 is dependent on the altitude of the SDS's panel.} Besides, all three phases of the Landing stage upload the AUH's operating status 1.5 times per minute which is same frequency as the Homing phase. It's worthy pointing out that a liner speed decision strategy, as obtained from Equation \ref{speed decision}, is utilized in both Landing1 and Landing2 phase. The objective of this strategy is to maintain the AUH in close proximity to the coordinate $(0,0)$, while also preventing the AUH from sliding past the origin caused by its significant inertia.
\begin{equation}
v=\begin{cases}
v_{tr}& \mathrm{if,} \ r >R_{o}\\
\frac{v_{tr}(r-R_i)}{(R_{o} -R_{i})} & \mathrm{if,} \ R_{o} \geqslant r >R_{i}\\
0 & \mathrm{if} ,\ r\leqslant R_{i}
\label{speed decision}
\end{cases}
\end{equation}
where, $v$ and $v_{tr}$ \sethlcolor{green}\hl{denotes} the desired speed and the presupposed transit speed, respectively. $r$ represents the distance between $(0,0)$ and the current position of the AUH. $R_{i}$ and $R_o$ refer to the inner radius and outer radius, respectively. By combining Equation \ref{speed decision} with its corresponding schematic diagram in Fig. \ref{FIG:3.6}, it is evident that the predefined outer radius ($R_o$) must be smaller than the effective illuminate radius ($R$). This requirement guarantees that the AUH remains within the visible range of illumination throughout the entire duration.

\begin{table}[width=1\linewidth,cols=3,pos=h]
\caption{The parameters of three phase in Landing.}\label{tbl1}
\begin{tabular*}{\tblwidth}{@{} LLLL@{} }
\toprule
 & Landing1   & Landing2   & Landing3     \\
\midrule
distance threshold(m) & 1        & 0.7        & --         \\
work altitude(m)      & 5          & 3.5        & 0.2          \\
yaw threshold(°)      & --       & 10         & 45           \\
pitch threshold(°)    & --       & --       & 5            \\
row threshold(°)      & --       & --       & 5            \\
depth threshold(m)    & --       & --       & 0.2          \\
transit\_speed(m/s)    & 0.3       & 0.3    & --           \\
outer\_speed(m/s)      & 0.2        & 0.2        & --          \\
outer\_radius(m)      & 1.5      & 1.5      & --            \\
inner\_radius(m)      & 0.3      & 0.3      & --            \\
navigation mode       & optical     & optical     & optical       \\
upload status         & Yes        & Yes        & Yes         \\
\bottomrule
\label{table 2}
\end{tabular*}
\end{table}

\begin{figure}[!h]
	\centering
\includegraphics[width=0.3\textwidth]{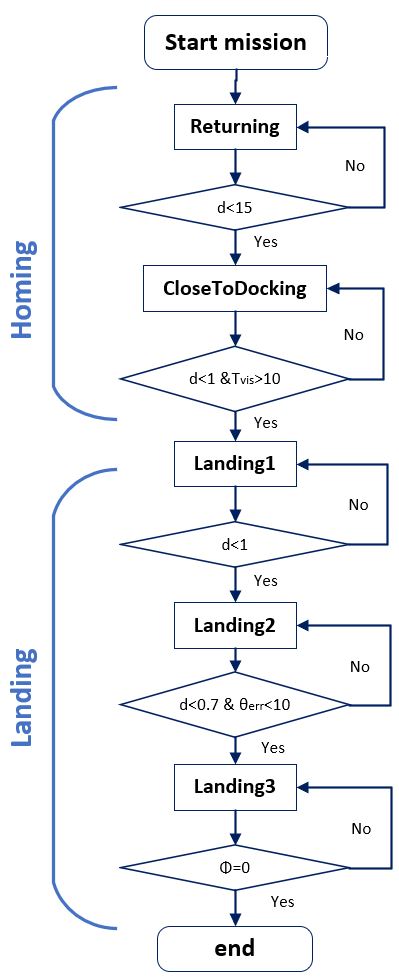}
	\caption{Flowchart of the whole docking procedure. $d$ refers the distance between the current position $(x,y)^T$ and the center of the SDS $(0,0)^T$. $T_{vis}$ \sethlcolor{green}\hl{denotes} the duration of entry into the effective illumination range. $\theta_{err}$ represents the difference of $\theta_d$ and $\theta_v$.}
	\label{FIG:3.4}
\end{figure}

\begin{figure}[!h]
	\centering
\includegraphics[width=0.48\textwidth]{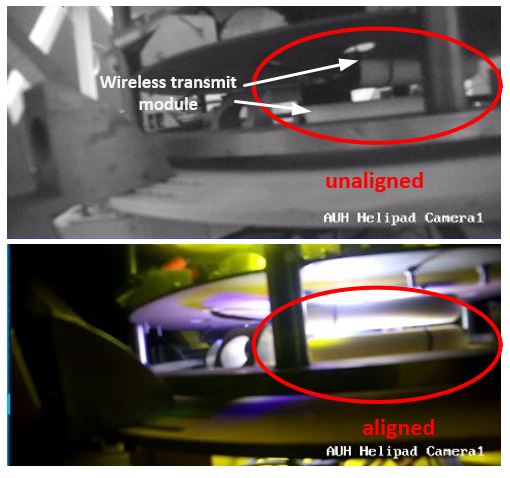}
	\caption{Wireless transmit module aligned.}
	\label{FIG:3.5}
\end{figure}

Landing3 is the final step in the docking procedure. After landing on SDS chassis, AUH will be locked by a steering gear mounted on the chassis for data transmission and wireless charging. However, when AUH is in Landing3 stage, several situations may arise. At first, the yaw angle may not be aligned properly (as shown in Fig.\ref{FIG:3.5}). Additionally, the AUH may be situated on the frame or seabed, or the chassis may interfere with the AUH. These situations can lead to issues such as the wireless module being too far away, affecting the rate of wireless transmission, or the steering gear losing its functionality and being unable to secure the AUH. To address these issues, a criterion is required to determine whether the AUH has successfully landed on the SDS. If the criterion is not met, the AUH will need to return to the Landing2 stage. Only when the conditions are met will the Landing3 stage be considered complete. 

\begin{equation}
\Phi  = f\left( \theta  \right) + f\left( \varphi  \right) + f\left( \phi  \right) + f\left( z \right)
\label{equ 5}
\end{equation}

\begin{equation}
f\left( \theta  \right) = \left\{ {\begin{array}{*{20}{c}}
0&{{\rm{if,}}\left| {\theta  - {\theta _d}} \right| \notin \left[ {0,{\theta _{thr}}} \right]}\\
1&{{\rm{if,}}\left| {\theta  - {\theta _d}} \right| \in \left[ {0,{\theta _{thr}}} \right]}
\end{array}} \right.
\label{equ 6}
\end{equation}

\begin{equation}
f\left( \varphi  \right) = \left\{ {\begin{array}{*{20}{c}}
0&{{\rm{if,}}\left| {\varphi  - {\varphi _d}} \right| \notin \left[ {0,{\varphi _{thr}}} \right]}\\
1&{{\rm{if,}}\left| {\varphi  - {\varphi _d}} \right| \in \left[ {0,{\varphi _{thr}}} \right]}
\end{array}} \right.
\label{equ 7}
\end{equation}

\begin{equation}
f\left( \phi  \right) = \left\{ {\begin{array}{*{20}{c}}
0&{{\rm{if,}}\left| {\phi  - {\phi _d}} \right| \notin \left[ {0,{\phi _{thr}}} \right]}\\
1&{{\rm{if,}}\left| {\phi  - {\phi _d}} \right| \in \left[ {0,{\phi _{thr}}} \right]}
\end{array}} \right.
\label{equ 8}
\end{equation}

\begin{equation}
f\left( z \right) = \left\{ {\begin{array}{*{20}{c}}
0&{{\rm{if,}}\left| {z - {z_d}} \right| \notin \left[ {0,{z_{thr}}} \right]}\\
1&{{\rm{if,}}\left| {z - {z_d}} \right| \in \left[ {0,{z_{thr}}} \right]}
\end{array}} \right.
\label{equ 9}
\end{equation}
where, $\Phi$ is the criterion mentioned above, and $\theta_{thr}$, $\varphi_{thr}$, $\phi_{thr}$ and $z_{thr}$ \sethlcolor{green}\hl{denote} the threshold of attitude and depth, respectively. In a conclusion, the flowchart outlining the phases in the whole docking procedure can be seen in Fig.\ref{FIG:3.4}.

\section{Experiments}\label{Experiments}
\subsection{Pool Experiment}
\begin{figure}[!h]
	\centering
\includegraphics[width=0.48\textwidth]{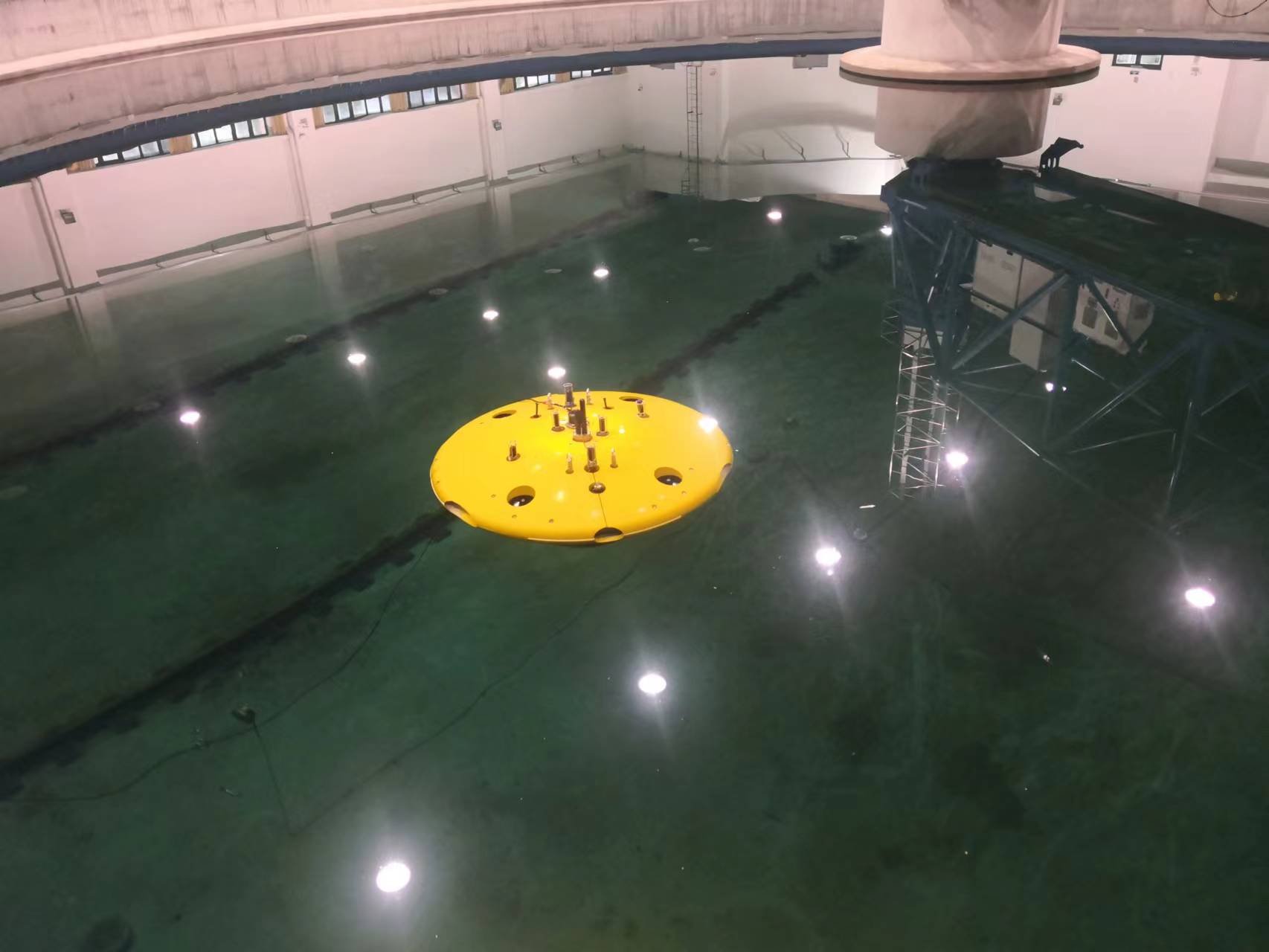}
	\caption{Pool experiments.}
	\label{FIG:4.7}
\end{figure}

The pool experiment on acoustic-inertial-optical guided docking of an AUH was carried out in the maneuvering pond at Zhejiang University, China(as shown in Fig.\ref{FIG:4.7}). The pond has a diameter of 45 meters and a water depth of 6 meters. The main objective of the experiment was to validate the feasibility and reliability of guidance method based on acoustic-inertial-optical principle, as well as control system of AUH, in a docking procedure.

\begin{figure}[!h]
	\centering
\includegraphics[width=0.48\textwidth]{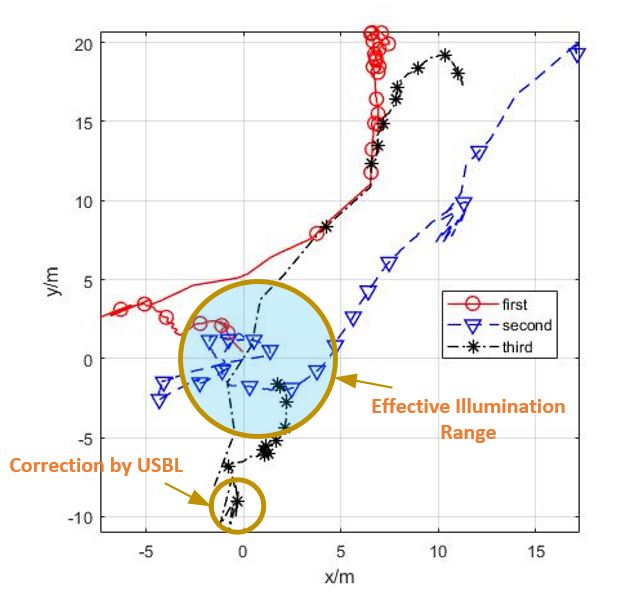}
	\caption{The plane trajectory from Returning to CloseToDocking phase.}
	\label{FIG:4.1}
\end{figure}

\begin{figure}[!h]
	\centering
\includegraphics[width=0.48\textwidth]{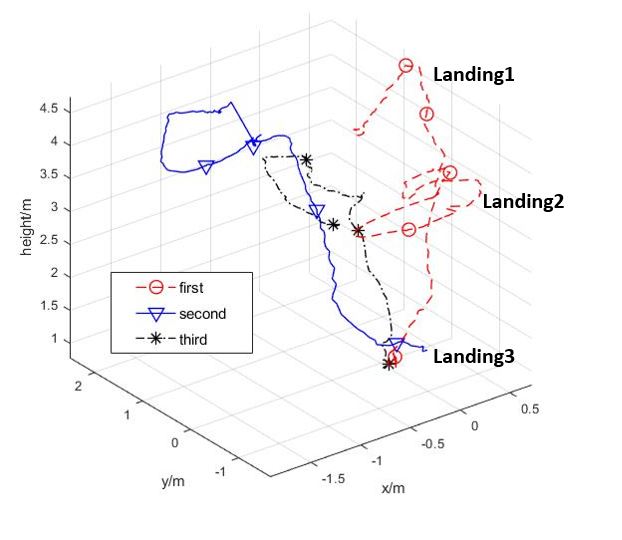}
	\caption{The 3-D trajectory in Landing stage.}
	\label{FIG:4.2}
\end{figure}

\begin{figure*}[!h]
	\centering
\includegraphics[width=0.9\textwidth]{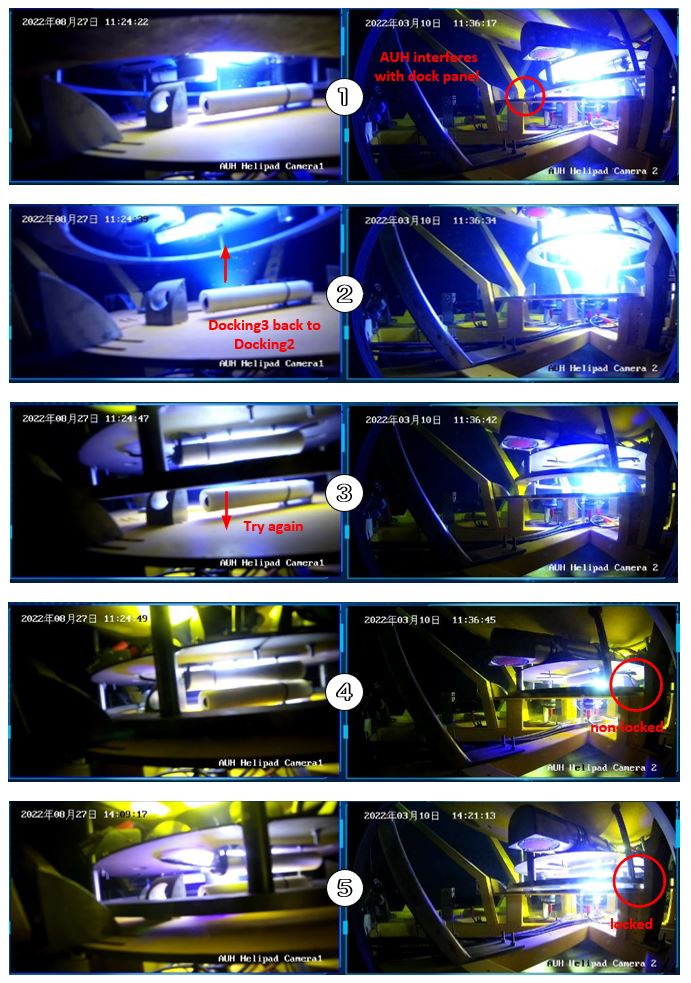}
	\caption{The landing moment of Landing3 phase. }
	\label{FIG:4.4}
\end{figure*}

Three representative trajectories were presented in Fig. \ref{FIG:4.1} and Fig. \ref{FIG:4.2}. In all three mission, the AUH successfully approached the SDS and enter the effective illumination range from a distance at least 20 meters. Although the second mission temporarily went out of the illumination range, the AUH was able to adjust its yaw angle and navigate back to the SDS using the acoustic-inertial navigation system. Obviously, AUH experienced three phase in Landing stage, for it hovered at three altitudes: 4.8 m , 3.2 m and 0.2 m, always near the point (0,0) in X-Y plane. The three curves in Fig. \ref{FIG:4.2} demonstrate that the AUH ultimately landed on the SDS panel. The horizontal coordinate data in Fig. \ref{FIG:4.1} is obtained from acoustic-inertial guidance based on USBL, DVL, and IMU. As a contrast, the horizontal coordinate data in Fig. \ref{FIG:4.2} is obtained from vision guidance based on monocular camera, while the altitude data is obtained from the depth altimeter. In Fig. \ref{FIG:4.4}, AUH initially encountered interference with the dock panel whose attitude is also quite different from the attitude of the SDS. As a result, the phase of AUH was transitioned from Landing3 to Landing2. Once distance and yaw angle met the specified thresholds, AUH resumed its descent, successfully landed on the panel. At this point,the operator remotely activated the steering gear and locked the AUH in place.

\subsection{Sea trial}
\begin{figure}[!h]
	\centering
\includegraphics[width=0.5\textwidth]{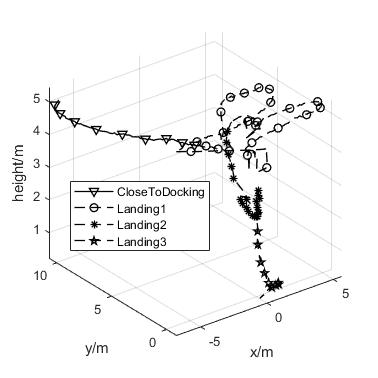}
	\caption{The 3-D trajectory of whole docking procedure in the sea trail. }
	\label{FIG:4.5}
\end{figure}

\begin{figure}[htp]
	\centering
\includegraphics[width=0.3\textwidth]{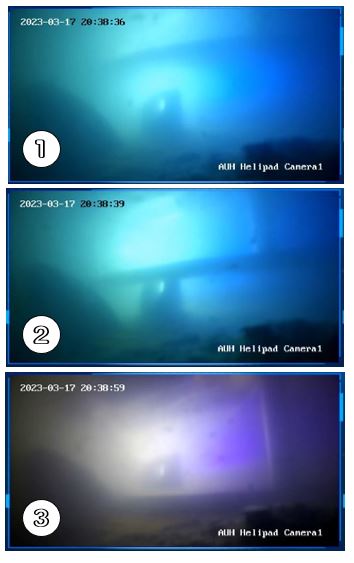}
	\caption{The docking moment in the sea trial. }
	\label{FIG:4.6}
\end{figure}

The sea trial was conducted in the South China Sea with the aim of validating the practicability and robustness of control system in a real sea environment. Fig. \ref{FIG:4.5} illustrates a representative curve from the CloseToDocking phase to the Landing3 phase. Similar to the pool experiments, the AUH curised back to the SDS from a distance of 12 meters. It then entered into Landing stage and performed three phase at various altitude: Landing1, Landing2, Landing3. Despite the challenging conditions in the sea trial, such as current disturbances and high turbidity sea water, the AUH successfully landed on the center of the SDS, and the whole docking procedure took about 13 minutes. (as shown in Fig.\ref{FIG:4.6})

\section{Conclusions}\label{conclusions}
A control system structure for AUH docking procedures based on acoustic-inertial-optical guidance is presented in this paper. In contrast to torpedo-type AUV, docking procedure of the AUH emphasis motion mobility in three-dimensional space, while conventional AUVs mainly focus on vertical motion. Building on previous work, this paper utilizes commercial multiple sensors fusion navigation technology based on IMU, DVL, USBL and monocular camera. At the software aspect, AUH is mainly comprised of four modules, namely Navigation and Communication, Motion Control, MOOSDB and HelmIvp. Under the management of HelmIvp, predefined behaviors generate desired values such as speed, altitude and attitude, which are then delivered to the motion control subsystem. The docking procedure is divided into two stages: Homing and Landing. In the Homing stage, the AUH returns to the SDS using acoustic-inertial guidance. Upon entering the effective illumination range, the AUH transitions to the Landing stage and lands on the center of the SDS using optical guidance. To counteract the influence of inertia and current disturbance, a segmented aligning strategy operating at various altitudes and a linear velocity decision are both adopted in Landing stage. As the distinctive design of the SDS, the AUH is required to dock onto it in a fixed orientation with precise attitude and altitude. To evaluate the successful docking, a particular criterion is proposed. The pool experiments demonstrated the feasibility of the control system, and subsequent sea trials validated its robustness and practicality in real sea conditions. 

Although AUH successfully docking onto the SDS, the procedure still takes considerable time and the success rate remained low. To address these problems, future work could involve implementing a motion controller with higher accuracy and faster response.

\section*{Acknowledgement}
This research was supported by the National Natural Science Foundation of China (Grant No. 52001279). The authors would like to thank Zhikun Wang and other Ocean College people at Zhejiang University for their inspiration and helping experiments.
 \bibliographystyle{cas-model2-names}
\bibliography{cas-dc}

\end{document}